\documentclass[conference]{IEEEtran}
\IEEEoverridecommandlockouts
\usepackage{cite}
\usepackage{amsmath,amssymb,amsfonts}
\usepackage{algorithmic}
\usepackage{graphicx}
\usepackage{textcomp}
\usepackage{xcolor}
\def\BibTeX{{\rm B\kern-.05em{\sc i\kern-.025em b}\kern-.08em
    T\kern-.1667em\lower.7ex\hbox{E}\kern-.125emX}}
\begin{document}

\title{Mask-Robust Face Verification for Online Learning via YOLOv5 and Residual Networks\\
\thanks{Corresponding author: Zhifeng Wang and Chunyan Zeng, Email: zfwang@ccnu.edu.cn, cyzeng@hbut.edu.cn}
}

\author{\IEEEauthorblockN{Zhifeng Wang}
	\IEEEauthorblockA{\textit{CCNU Wollongong Joint Institute} \\
		\textit{Central China Normal University}\\
		Wuhan 430079, China \\
		zfwang@ccnu.edu.cn}
	\and
	\IEEEauthorblockN{Minghui Wang}
	\IEEEauthorblockA{\textit{CCNU Wollongong Joint Institute} \\
		\textit{Central China Normal University}\\
		Wuhan 430079, China \\
		mw402@uowmail.edu.au}
	\and
	\IEEEauthorblockN{Chunyan Zeng}
	\IEEEauthorblockA{\textit{School of Electrical and Electronic Engineering} \\
		\textit{Hubei University of Technology}\\
		Wuhan 430068, China \\
		cyzeng@hbut.edu.cn}
	\and
	\IEEEauthorblockN{Jialong Yao}
	\IEEEauthorblockA{\textit{CCNU Wollongong Joint Institute} \\
		\textit{Central China Normal University}\\
		Wuhan 430079, China \\
		jy123@uowmail.edu.au}
	\and
	\IEEEauthorblockN{Yang Yang}
	\IEEEauthorblockA{\textit{CCNU Wollongong Joint Institute} \\
		\textit{Central China Normal University}\\
		Wuhan 430079, China \\
		univeryang@ccnu.edu.cn}
	\and
	\IEEEauthorblockN{Hongmin Xu}
	\IEEEauthorblockA{\textit{Faculty of Artificial Intelligence in Education} \\
		\textit{Central China Normal University}\\
		Wuhan 430079, China \\
		xhm@ccnu.edu.cn}
}

\maketitle

\IEEEpubidadjcol

\begin{abstract}
In the contemporary landscape, the fusion of information technology and the rapid advancement of artificial intelligence have ushered school education into a transformative phase characterized by digitization and heightened intelligence. Concurrently, the global paradigm shift caused by the Covid-19 pandemic has catalyzed the evolution of e-learning, accentuating its significance. Amidst these developments, one pivotal facet of the online education paradigm that warrants attention is the authentication of identities within the digital learning sphere.
Within this context, our study delves into a solution for online learning authentication, utilizing an enhanced convolutional neural network architecture, specifically the residual network model. By harnessing the power of deep learning, this technological approach aims to galvanize the ongoing progress of online education, while concurrently bolstering its security and stability. Such fortification is imperative in enabling online education to seamlessly align with the swift evolution of the educational landscape.
This paper's focal proposition involves the deployment of the YOLOv5 network, meticulously trained on our proprietary dataset. This network is tasked with identifying individuals' faces culled from images captured by students' open online cameras. The resultant facial information is then channeled into the residual network to extract intricate features at a deeper level. Subsequently, a comparative analysis of Euclidean distances against students' face databases is performed, effectively ascertaining the identity of each student. It's noteworthy that the innovative approach surpasses conventional YOLOv5 performance by offering specialized face detection, a critical function particularly relevant in the current pandemic milieu where face masks are commonplace. Furthermore, our methodology seamlessly extends the traditional YOLOv5 network's capabilities.
\end{abstract}


\section{Introduction}
In recent years, the adoption of various teaching methods, such as cloud classrooms and online learning, has been accelerated by the Covid-19 pandemic \cite{Wang2025,Shi2026,Wang2025b,Dong2025,Wang2025d,Chen2025b,Wang2024m,Chen2024e,Wang2023v,Ma2023b,Wang2022as}. These methods have introduced new challenges \cite{Wang2023j,Liao2024,Wang2024p,Min2019,Wang2023g,Chen2024g}, as traditional face-to-face teaching is limited and conventional attendance tracking methods are often insufficient in online settings \cite{Li2023f,Lyu2022}. Consequently, the need for AI-assisted student information verification has grown in prominence \cite{Adedoyin2023,Wang2025f,Li2026a,Wang2025e,Li2023i,Wang2024b,Li2023g,Wang2024s,Li2023f,Wang2023j,Dong2023,Wang2023d,Lyu2022}. The domain of computer vision, especially deep learning-based techniques, has rapidly evolved and found applications across diverse fields \cite{Zheng2025,Zeng2025,Chen2025a,Zeng2024g,Chen2025,Zeng2024h,Zheng2024,Zeng2024b,Wang2023f,Zeng2024f,Chen2023b,Zeng2024c,Wang2022t,Zeng2024d,Wang2021m,Zeng2024,Wang2020h,Zeng2024a,Wang2018a,Zeng2023a,Wang2015b,Zeng2023,Zhu2013,Zeng2022a,Wang2011,Zeng2021a,Wang2011a,Zeng2021b,Zeng2020,Zeng2018}. Leveraging these advancements, we can develop target detection and face recognition algorithms that accurately record students' attendance in online education settings.

Within the realm of computer vision, three main target detection algorithms are prominent: Faster R-CNN \cite{Girshick2015a}, SSD \cite{Liu2016b}, and YOLO \cite{Redmon2016,Redmon2017,Redmon2018,Bochkovskiy2020,Wang2022at,Wang2023k}. Faster R-CNN operates through a two-stage process involving Region Proposal Network (RPN) and Fast R-CNN. This approach excels in accuracy, particularly in complex scenes and multi-category detection tasks, but at the cost of speed \cite{Girshick2015a}. On the other hand, YOLO \cite{Li2023} and SSD \cite{Liu2016b}, belonging to the one-stage category, treat target detection as a regression challenge, accomplishing both target localization and classification in a single step. These methods offer both high detection accuracy and faster processing speed \cite{Wang2022at}.

In the field of face recognition, algorithms for feature extraction play a critical role in converting facial images into distinct feature vectors for matching and identification. Existing methods like Local Binary Pattern \cite{Heikkila2009} and Histogram of Oriented Gradients (HOG) \cite{Deniz2011} have been employed. Concurrently, the rise of deep learning has ushered in convolutional neural networks (CNNs) \cite{Wang2023a,Zeng2023c}, enabling the automatic extraction of intricate and abstract feature representations, leading to significant performance advancements in face recognition tasks.

In summary, our contributions encompass three key aspects:
\begin{itemize}
	\item Leveraging convolutional neural networks, particularly the residual network architecture, we have developed an efficient and effective face recognition system tailored for application within cloud classrooms and online learning platforms.
	\item In light of the ongoing pandemic, we have implemented a masked face recognition capability, allowing accurate recognition of individuals even when wearing masks.
	\item Employing a custom face dataset for training, the YOLOv5 model used for target detection achieved precision of 0.83494 and recall of 0.81308. Furthermore, the optimized version of the ResNet-34 face recognition model attained a mean error of 0.993833 with a standard deviation of 0.00272732 on the LFW benchmark.
	
\end{itemize}

The subsequent sections are structured as follows: Section \ref{rw} outlines the methodologies employed, encompassing target detection, feature extraction, database construction, and front-end development. Section \ref{Method} elaborates on the core principles underlying our framework for target detection and feature extraction, while Section \ref{Exp} succinctly summarizes experimental outcomes, supported by visualized training graphs. Concluding our work, we reflect on our achievements and chart potential avenues for future research in Section \ref{Con}.

\section{Related Work} \label{rw}

\subsection{Object Detection Using YOLO}

The YOLO (You Only Look Once) algorithm has emerged as a prominent single-stage target detection model, especially favored for its exceptional speed and accuracy on mobile devices compared to its two-stage counterparts \cite{Redmon2016}. Joseph Redmon first introduced YOLO in 2015, aiming to achieve real-time target detection with sustained precision. The evolution of YOLO algorithms comprises several iterations \cite{Terven2023}.

YOLOv1, the initial version, frames target detection as a regression problem, predicting both location and category via a single forward propagation. It divides the input image into grid cells and predicts bounding boxes and category probabilities on each cell \cite{Redmon2016}. However, YOLOv1 is marred by issues of imprecise localization and challenges in detecting small targets. Subsequent iterations, such as YOLOv2, employ Darknet-19 as the base network, incorporating additional convolutional layers to enhance feature extraction \cite{Redmon2017}. YOLOv3 further refines the model's performance \cite{Redmon2018}, incorporating increased convolutional layers and Residual Blocks to construct a more intricate network. YOLOv4 advances the paradigm with a larger architecture, adopting CSPDarknet53 as the backbone network and introducing new target detection loss functions alongside technical optimizations, culminating in remarkable accuracy and speed improvements \cite{Bochkovskiy2020}.

This paper employs the streamlined CSPDarknet53 network architecture, delivering superior detection performance. It also integrates data augmentation techniques and supports half-precision training, facilitating accurate student face frame detection.

\subsection{Feature Extraction and Comparison}

The domain of computer vision has experienced significant advancements through deep learning-based feature extraction methods, with deep neural network (DNN) based techniques gaining prominence \cite{Wang2023v,Zeng2025a,Wang2023l,Zeng2024e,Wang2022at,Zeng2023c,Wang2025g,Li2023h,Zeng2022,Wang2022ac,Zeng2022b,Wang2021,Zeng2021c,Tian2018,Zeng2020a,Min2018,Wang2017,Wang2015a}. Within DNN-based feature extraction, pretrained CNN models \cite{Wang2023l} are frequently employed, with their early or intermediate convolutional layers utilized as feature extractors. These layers yield Feature Maps, capturing abstract image characteristics at diverse levels \cite{Wang2023g}. Extracting these feature maps yields enriched, abstract image representations, bolstering the efficacy of tasks like image recognition, detection, and classification. This development encompasses concepts like automatic feature representation learning, leveraging pre-trained models for mitigating data and computational constraints, and integrating methods like feature fusion and multi-task learning to enhance image recognition and analysis performance \cite{Zeng2022}.

Our feature extraction methodology centers on the ResNet technique, a deep convolutional neural network architecture devised by Kaiming He. ResNet introduces the Residual Block, simplifying the training of exceedingly deep networks by mitigating issues like gradient vanishing and explosion encountered in traditional deep networks. We employ an improved ResNet to extract high-dimensional feature vectors corresponding to student faces for precise identity determination.

\subsection{Online Learning Authentication System Construction}

PyQt, a Python binding to the Qt application framework, facilitates cross-platform graphical interface application development through Python. Developers leverage PyQt's access to Qt's functionalities to craft feature-rich GUI applications using Python. PyQt offers a comprehensive array of classes and functions for creating GUI elements such as windows, buttons, labels, and text boxes, enabling interaction through the signals and slots mechanism. Its style and theme support empowers developers to customize interface appearances easily. In the context of constructing a student-assisted monitoring system, PyQt facilitates the creation of intuitive interfaces for students and teachers, facilitating student authentication. Simultaneously, we build a student face database storing high-dimensional feature vectors, including student ID, name, and facial features, to serve as the foundation for student authentication. ResNet network-extracted features enable identity comparisons, culminating in effective student authentication.

\begin{figure*}[htbp]
	\centering
	\includegraphics[width=1\textwidth]{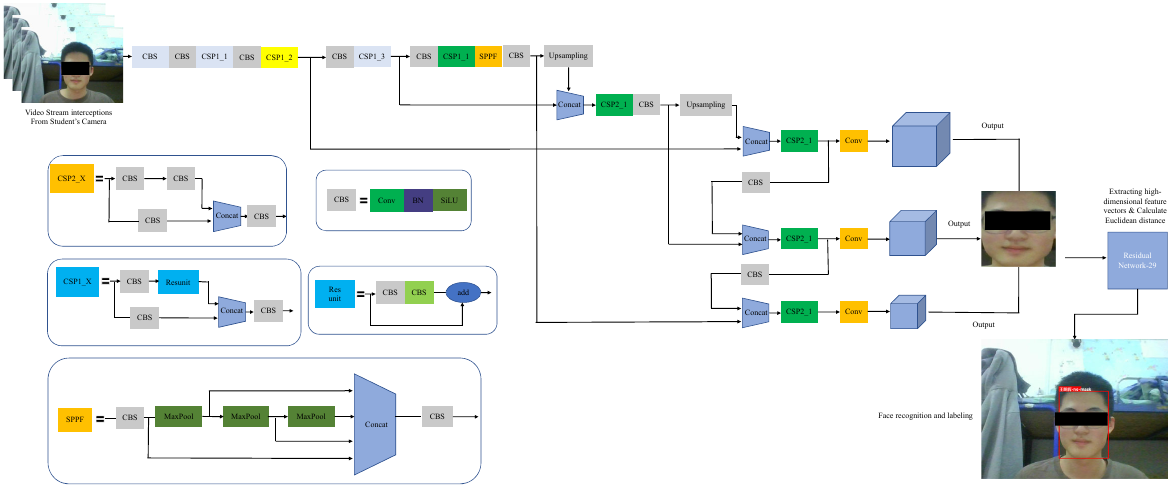}
	\caption{The framework of student authentication system. It consists of a head section for preprocessing, a backbone section and a SPPF section for locating face coordinates, and a ResNet-29 section for extracting high-dimensional face features.}
	\label{fig1}
\end{figure*}

\section{Proposed Method} \label{Method}
This section outlines the method proposed in this paper, encompassing the utilization of YOLOv5 for target face detection, the extraction of corresponding deep features with the assistance of ResNet, and the principles underlying each component, as shown in Fig. \ref{fig1}.

\subsection{Network Design for Student Face Recognition}

The approach begins with employing YOLOv5 to detect target faces, distinguishing between two categories: faces wearing masks and faces without masks. This step involves labeling the coordinates of the located faces. Subsequently, ResNet is employed to extract the relevant deep features, which are then matched against the features stored in the database for comparison. The following paragraphs elucidate each aspect of this process.

The Network Design for Student Face Recognition phase commences by extracting a single frame of a student's face from the video stream captured by the e-learning system's camera. The structure of the student authentication system converts this frame into a standardized format of 640$\times$640$\times$3 for uniform processing. This standardized frame is then fed into a Backbone network for feature extraction. Our framework employs the CSPDarknet53 architecture as the Backbone. This architecture is an advanced iteration of the Darknet design, incorporating Cross Stage Partial (CSP) connections to enhance feature extraction efficiency and performance. The core innovation of the CSP technique involves bifurcating the feature map into two segments. One segment traverses a series of convolutional layers, while the other bypasses selected convolutional layers. These two segments are subsequently combined, enabling cross-layer information interaction and bolstering feature characterization capabilities. This feature allows the network to harness both shallow and deep features at each stage, enhancing object detection capabilities across various scales. The CSP technology is applied to both the Backbone and Neck components, each composed of convolutional layers and batch normalization. The components are finalized with a SiLu activation function:

\begin{equation}
SiLu(x)=x*sigmoid(x)
\end{equation}

Following feature extraction, the Neck layer processes the multi-layered feature maps delivered by the Backbone network. Through a fusion of deep and shallow information via Feature Pyramid Network (FPN) and PAN, this layer generates the corresponding detection categories, transmitting them to the subsequent Head layer. Within the Head layer, channel expansion accommodates different scales of feature maps obtained from the Neck, executed via 1$\times$1 convolutions. This expansion augments the dimensions of features to encompass (number of categories + 5) multiplied by the number of channels per detection. The "5" components encapsulate the horizontal and vertical coordinates of the prediction frame's center, width, height, and confidence level. Confidence level values indicate prediction frame credibility, with higher values signifying a greater likelihood of containing the target.

\begin{figure}[h]
    \centering
    \includegraphics[width=0.49\textwidth]{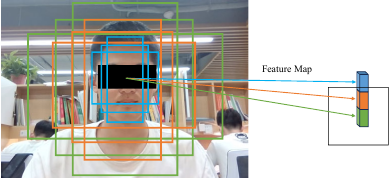}
    \caption{Feature maps of varying sizes.}
    \label{fig2}
\end{figure}

The Head layer integrates three detection layers, aligned with the diverse sizes of feature maps derived from the Neck. These layers segment grids on the maps based on size, configuring three anchors with varied aspect ratios for each grid. These anchors underpin target prediction and regression. The extended channel dimensions facilitate encoding positional and categorization data, elaborated in Fig. \ref{fig2}.

In the final stage of the detection, a non-maximal suppression (NMS) algorithm is employed to eliminate overlapping bounding boxes and identify the optimal bounding box. The NMS algorithm assesses the confidence scores of bounding boxes along with the extent of overlap, subsequently eliminating redundant bounding boxes. To evaluate the precision of the prediction box, the forecasted information is compared against the ground truth data, directing the model's iterative convergence.

The loss function we utilize in this context is the Complete Intersection over Union (CIoU). This loss function quantifies the disparity between predicted and true information. When the predicted information approximates the actual data, the loss function's value diminishes. The traditional metric for assessing overlap between the predicted and actual frames is Intersection over Union (IoU), expressed as:

\begin{equation}
IoU~ = ~\frac{Prediction Bound\cap Real Bound}{Prediction Bound\cup Real Bound}
\end{equation}

However, IoU has limitations. When the prediction frame does not intersect with the real frame, IoU cannot gauge the spatial separation between the two frames, resulting in a loss of IoU accuracy. This leads to an IoU loss of 0, impacting gradient backpropagation and ultimately undermining training effectiveness. Furthermore, IoU struggles to precisely measure the magnitude of overlap between the prediction and real frames.

To surmount these limitations, the CIoU was introduced as a refinement, as implemented in YOLOv5. This modification improves IoU by considering additional factors. The CIoU is formulated as:

\begin{equation}
CIoU~ = ~IoU-(\frac{\rho^2 (b,b^{gt})}{c^2}+\alpha \upsilon)
\end{equation}
Here, $\alpha$ functions as a weighting parameter, and $\upsilon$ gauges the aspect ratio's consistency. The specific definitions are as follows:

\begin{equation}
\alpha = ~\frac{\upsilon}{(1-IoU)+\upsilon}
\end{equation}

\begin{equation}
\upsilon= ~\frac{4}{\pi^2}(arctan\frac{w^{gt}}{h^{gt}}-arctan\frac{w}{h})^2
\end{equation}

In essence, the CIoU refines the loss function, rectifying IoU's limitations and enhancing the model's accuracy and training stability. This demonstrates the continual advancement of techniques in pursuit of more effective and reliable object detection models.

\subsection{Feature Extraction Process for Student Faces}

Feature extraction is a pivotal process that involves extracting meaningful and informative feature representations from raw data, enabling the capture of crucial data characteristics to enhance subsequent tasks' effectiveness.

ResNet (Residual Neural Network) emerges as an advantageous choice for feature extraction, primarily due to its deep network architecture and the incorporation of a residual connectivity mechanism. ResNet, known for its depth, can accommodate numerous network layers, granting it the capacity to grasp intricate and abstract feature representations. This capacity is especially advantageous in domains like image analysis, speech recognition, and natural language processing. In computer vision tasks, such as image classification and target detection, deeper networks are adept at assimilating richer image details, thereby enhancing model performance. Traditional deep networks confront the challenge of gradient vanishing or explosion as the number of layers increases, impeding network training. ResNet tackles this hurdle via the introduction of residual connections. Within each residual block, the input passes through a residual function before being added to the original input. This strategy preserves the original input's information and facilitates gradient propagation. The employment of residual connections augments training stability, addresses vanishing gradient concerns, and concurrently reduces parameter count. The residual block function in our structure is illustrated in Fig. \ref{fig3}:

\begin{figure}[h]
    \centering
    \includegraphics[width=0.48\textwidth]{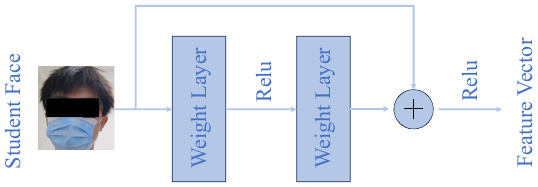}
    \caption{Feature extraction structure.}
    \label{fig3}
\end{figure}

Every network layer introduces a residual function signifying the difference between the network layer's output and the input. Concretely, when the input traverses the residual block, the output is expressed as:

\begin{equation}
H(x)=F(x)+x
\end{equation}

Where $x$ signifies the input and $F(x)$ represents the residual function. This function quantifies the deviation between the network layer output and the input. By amalgamating the output of the residual function with the input, the network layer yields its final output. This strategy transforms the learning objective from comprehending the original mapping $H(x)$ to mastering the residual function $F(x)$. Learning residuals proves more straightforward than mastering the complete mapping, thereby enabling ResNet to facilitate the construction of deeper networks, leading to improved feature extraction. Capitalizing on the advantages of ResNet, we opt for a pre-trained model constructed upon ResNet-34 for student face feature extraction. This ResNet model, curated by Davis E. King, consists of 29 convolutional layers. Essentially, it is a customized variant of the ResNet-34 network elucidated in the paper "Deep Residual Learning for Image Recognition" by He, Zhang, Ren, and Sun. It involves the removal of certain layers and a 50\% reduction in the number of filters per layer. The model was trained from scratch using an expansive dataset containing approximately 3 million face images, amalgamated from various sources.

Subsequent to forwarding face coordinates through the aforementioned residual network, the process yields 128D high-dimensional feature vectors in corresponding coordinates. These vectors are then matched against the facial information of students stored in the database. This entails computing the Euclidean distance between the vectors, culminating in the provision of the most fitting facial information match.

\section{Experimental Results and Analysis} \label{Exp}
This section presents an overview of our experimental configuration, training dataset, and outlines the outcomes of the final experiment. The results are also visualized using graphical representations.

\subsection{Experimental Environment}
The experiment was conducted using the PyTorch 1.10.2 framework with CUDA version 11.3. The relevant computer vision libraries included OpenCV 4.4.5.64, PyQt 5.15.4, and Tensorboard 2.6.0. The experiment was carried out on a Windows 11 platform, utilizing Python 3.8. The hardware setup comprised an NVIDIA GeForce 4060 graphics card, 16 GB of RAM, and training was performed using a 512 GB solid-state drive for data caching.

\subsection{Dataset Collection and Training}
Our dataset comprises 2000 facial images sourced from diverse scenarios on the web. This collection encompasses both masked and unmasked faces, with 1000 instances for each category, resulting in a balanced ratio of 1:1. These images were obtained from video streams, webcasts, and news media streams to mirror the imagery transmitted by the online learning system for enhanced relevance. The image dataset was meticulously annotated using "labelimg" software to mark the faces. The dataset was divided into training and testing sets in an 8:2 ratio to facilitate comprehensive evaluation.

\subsection{Analysis of Experimental Results}
In the evaluation of the model's performance, precision, recall, and mAP (mean average precision) are key metrics utilized to gauge the efficacy of the object detection system. The results are visualized to provide a clear understanding of the model's performance.

The precision and recall curves are depicted in Fig. \ref{fig4} and \ref{fig5}, respectively:

\begin{figure}[h]
    \centering
    \includegraphics[width=0.5\textwidth]{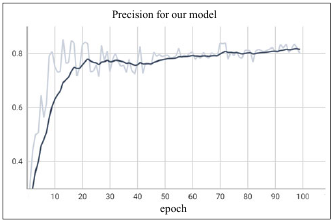}
    \caption{Precision for YOLOv5s.}
    \label{fig4}
\end{figure}
 
\begin{figure}[h]
    \centering
    \includegraphics[width=0.5\textwidth]{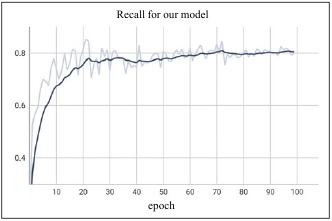}
    \caption{Recall for YOLOv5s.}
    \label{fig5}
\end{figure}

Table \ref{tab1} presents the model's performance metrics over different epochs:
\begin{table}[h]
\centering
\caption{The performance of proposed model.\label{tab1}}
\begin{tabular}{ccc}
\hline
\multicolumn{1}{l}{epoch} & precision & recall \\ \hline
0                         & 0.098143            & 0.12642        \\
10                        & 0.61519           & 0.67598         \\
20                        & 0.71638            & 0.71542        \\
30                        & 0.77438           & 0.78529        \\
40                        & 0.7364           & 0.74315        \\
50                        & 0.79626           & 0.79557        \\
60                        & 0.78905           & 0.78038        \\
70                        & 0.79472           & 0.8131        \\
80                        & 0.78097           & 0.78096        \\
90                        & 0.8157           & 0.81369        \\
100                       & 0.82104           & 0.7915         \\ \hline
\end{tabular}
\end{table}

The key metrics are computed as follows:
\begin{equation}
precision=\frac{TP}{TP+FP}
\end{equation}

\begin{equation}
Recall=\frac{TP}{TP+FN}
\end{equation}

\begin{equation}
mAP=\frac{\sum_{j=0}^{n}AP(j) }{n}
\end{equation}

In the context of our target detection evaluation, the terminology "Total Positive Detections (TP)" signifies the count of accurately identified targets by the model, aligning with instances where the model correctly predicts a target that aligns with the actual target. On the other hand, "Total Number of Negative Detections (TN)" refers to the quantity of non-target areas correctly identified and excluded by the model, corresponding to instances where the model effectively refrains from detecting a target in the background region. "False Negative (FN)" pertains to cases in which the model fails to accurately detect a target that is genuinely present. The metric "Average Precision (AP)" assumes significance in this target detection endeavor.

\begin{figure}[h]
    \centering
    \includegraphics[width=0.49\textwidth]{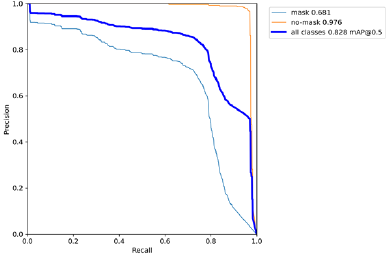}
    \caption{PRcurve for YOLOv5s.}
    \label{fig6}
\end{figure}

Precision and Recall constitute pivotal performance indicators within target detection tasks. Typically computed at varying thresholds, these metrics may yield multiple sets of precision and recall values. The average precision for the jth category, denoted as "AP(j)," can be established by computing the area under each curve within the precision-recall curves (P-R curves) plot, as shown in Fig. \ref{fig6}.

\begin{figure}[h]
    \centering
    \includegraphics[width=0.5\textwidth]{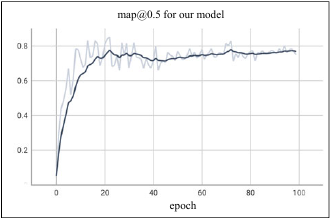}
    \caption{map@0.5 for YOLOv5s.}
    \label{fig7}
\end{figure}

Following the derivation of AP(j), the mean Average Precision (mAP) value can be calculated utilizing a specific formula. This evaluation metric amalgamates accuracy considerations across diverse categories and confidence thresholds, providing an average value encompassing all categories. The subsequent figure visually illustrates the model's performance in terms of the mean Average Precision value during the face detection phase as shown in Fig. \ref{fig7}.

The infographic depicts a face, representing the absence of a mask, while a mask symbolizes the presence of a mask on the face. To summarize, this experiment yielded relatively favorable results, which are then passed on to the subsequent ResNet model for face feature extraction. In comparison to the original YOLOv5 model, this iteration focuses more intently on extracting facial features, a refinement achieved through specialized dataset training. This model finds practical application in online learning assistance systems for students and holds potential for implementation in online learning assistance certification systems as well.

\begin{figure}[h]
    \centering
    \includegraphics[width=0.5\textwidth]{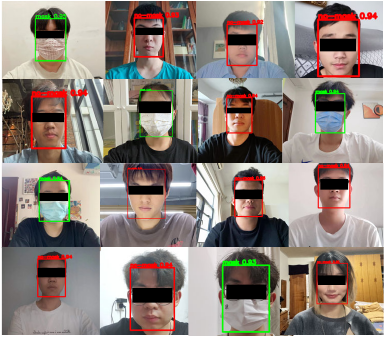}
    \caption{Student face detection.}
    \label{fig8}
\end{figure}

As illustrated in Fig. \ref{fig8}, this image exemplifies the detection of collected student face images. The rectangular boxes within each image demarcate detected student faces and categorize them based on mask-wearing status. Boxes labeled 'mask' denote student faces with masks, while those labeled 'no-mask' indicate student faces without masks.

\begin{figure}[htbp]
\centering
\begin{minipage}[t]{0.2\textwidth}
\centering
\includegraphics[width=3.25cm]{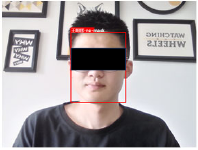}
\caption{Student without mask. \label{fig9}}
\end{minipage}
\hfill
\vspace{0.3cm}
\begin{minipage}[t]{0.2\textwidth}
\centering
\includegraphics[width=3.25cm]{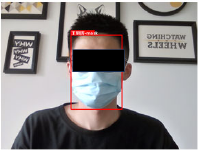}
\caption{Student with mask. \label{fig10}}
\end{minipage}
\end{figure}

Upon assessing the target detection function, our focus shifts towards evaluating the feature extraction module. For this purpose, we employ a pre-trained Residual Network (ResNet) model based on the ResNet-34 implementation designed by Davis King. This model achieves a mean error of 0.993833 with a standard deviation of 0.00272732 on the Labeled Faces in the Wild (LFW) benchmark dataset. The subsequent two images vividly showcase the authentication of student identity within our online learning system, as shown in Fig. \ref{fig9} and \ref{fig10}.

\section{Conclusion} \label{Con}
In this paper, we have developed an online learning authentication system that leverages a convolutional neural network architecture for the purpose of recognizing the faces of online students. Our approach involved training a dedicated YOLOv5 model using a self-constructed training dataset. This model excels in facial recognition and possesses the ability to discern whether a given face is adorned with a mask. Moreover, we achieved the essential task of feature extraction through a Residual Network (ResNet), facilitating the recognition of pertinent facial attributes. By employing these extracted facial features, we have successfully established a framework for verifying the identity of students.
The results of our comprehensive experiments substantiate the effectiveness of our proposed assisted authentication method. The YOLOv5 model's adeptness in face detection, coupled with the ResNet's proficiency in feature extraction, collectively contribute to a robust and accurate authentication system. The integration of these technologies showcases promising potential for ensuring the integrity and security of online learning environments. By fostering precise student identification, our system bolsters the credibility of online learning platforms and aids in preventing unauthorized access.
%The ongoing evolution of online education necessitates innovative solutions that maintain the quality and legitimacy of the learning process. Our approach, as demonstrated through empirical evaluation, signifies a notable stride in achieving this goal. Through the judicious fusion of advanced neural network architectures, we have addressed the challenges of student authentication in online learning contexts. While the current study showcases promising results, avenues for further exploration and refinement remain open, such as investigating the system's performance with larger and more diverse datasets, enhancing real-time processing capabilities, and adapting to evolving security paradigms.

\bibliographystyle{IEEEtran} 
\bibliography{Ref,Citations}

\end{document}